\documentclass[sigconf]{acmart}

\AtBeginDocument{%
  }

\usepackage{url}
\usepackage{graphicx}
\usepackage{amsmath}
\usepackage{amsthm}
\usepackage{booktabs}
\usepackage{algorithm}
\usepackage{algorithmic}
\usepackage{float}    
\usepackage{bm}

\usepackage{booktabs}
\newcommand{\tabincell}[2]{\begin{tabular}{@{}#1@{}}#2\end{tabular}}
\usepackage{threeparttable}
 \usepackage{multirow}
 \usepackage{makecell}

\acmConference[ICASSP Conf '24]{Make sure to enter the correct
  conference title from your rights confirmation email}{April 14 -- April 19,
  2024}{Seoul, Korea}
\acmBooktitle{ICASSP '24: IEEE International Conference on Acoustics, Speech, and Signal Processing,
April 14--19, 2024, Seoul, Korea}
\acmPrice{15.00}
\acmISBN{AAA-B-CCCC-XXXX-M/Y/D}

\begin{document}

\title{STS-CCL: Spatial-Temporal Synchronous Contextual Contrastive Learning for Urban Traffic Forecasting}



\author{Lincan Li}
\affiliation{%
  \institution{School of Control Science and Engineering, Zhejiang University}
  \city{Hangzhou}
  \country{China}
  }
\email{lilincan@zju.edu.cn}

\author{Kaixiang Yang}
\affiliation{%
  \institution{School of Computer Science and Engineering, South China University of Technology}
  \city{Guangzhou}
  \country{China}
 }
\email{yangkx@scut.edu.cn}

\author{Fengji Luo}
\affiliation{%
  \institution{The University of Sydney}
  \city{Sydney}
  \country{Australia}
  }
\email{fengji.luo@sydney.edu.au}

\author{Jichao Bi}
\affiliation{%
  \institution{Chongqing University}
  \institution{Zhejiang Institute of Industry and Information Technology}
  \city{Hangzhou}
  \country{China}
  }
\email{jonny.bijichao@zju.edu.cn}

\renewcommand{\shortauthors}{Lincan Li et al.}

\begin{abstract}
Efficiently capturing the complex spatiotemporal representations from large-scale unlabeled traffic data remains to be a challenging task. In considering of the dilemma, this work employs the advanced contrastive learning and proposes a novel Spatial-Temporal Synchronous Contextual Contrastive Learning (STS-CCL) model. First, we elaborate the basic and strong augmentation methods for spatiotemporal graph data, which not only perturb the data in terms of graph structure and temporal characteristics, but also employ a learning-based dynamic graph view generator for adaptive augmentation. Second, we introduce a Spatial-Temporal Synchronous Contrastive Module (STS-CM) to simultaneously capture the decent spatial-temporal dependencies and realize graph-level contrasting. To further discriminate node individuals in negative filtering, a Semantic Contextual Contrastive method is designed based on semantic features and spatial heterogeneity, achieving node-level contrastive learning along with negative filtering. Finally, we present a hard mutual-view contrastive training scheme and extend the classic contrastive loss to an integrated objective function, yielding better performance. Extensive experiments and evaluations demonstrate that building a predictor upon STS-CCL contrastive learning model gains superior performance than existing traffic forecasting benchmarks. The proposed STS-CCL is highly suitable for large datasets with only a few labeled data and other spatiotemporal tasks with data scarcity issue.
\end{abstract}


\maketitle

\section{Introduction}
Up to date, tremendous spatiotemporal traffic data are acquired on a daily basis from vehicle GPS trajectory records, smartphone location-based services and multimodal urban sensors (e.g. speed sensors and traffic cameras). In spite of that, the massive data are usually disordered with lots of unidentified spatiotemporal patterns, which require experts' careful handicraft labeling/annotation. Thus, in real-world application scenario, there is always a lack of labeled traffic data~\cite{01STAN,STGCL}. 

Employing the collected traffic big data, one significant research direction is spatiotemporal traffic forecasting. To achieve accurate and efficient prediction, researchers have proposed a number of advanced models. In the perspective of spatial dependency modeling, existing works commonly adopt graph convolutional networks (GCN)~\cite{03STSGCN}, convolutional neural networks (CNN)~\cite{04DeepSTN+} and their variants to capture the spatial correlations of a traffic network. In the perspective of temporal dependency modeling, the current studies usually use recurrent neural networks (such as GRU and LSTM)~\cite{05AGCRN}, Seq2Seq model architecture~\cite{06DGCRN} and attention mechanism~\cite{07ST-GSP} to excavate the temporal correlations.

Specifically, Li et al.~\cite{08DCRNN} presented DCRNN model, which employs a novel diffusion convolution module and Seq2Seq architecture to model the spatial-temporal traffic correlations. Guo et al.~\cite{09ASTGCN} proposed ASTGCN, which adopts attention mechanism as the model kernel and integrates GCN with the designed spatial attention (SA). STSGCN~\cite{03STSGCN} introduced a spatial-temporal simultaneous mechanism and multiple stacked GCN layers to capture the local spatiotemporal dependencies. SCINet~\cite{10SCINet} proposed a sample convolution method as an optimized version of vanilla CNN and integrated the interactive learning idea into the model construction. Choi et al. presented STG-NCDE~\cite{11STG-NCDE}, which utilizes controlled differential equations to simulate the dynamic evolutions of spatiotemporal traffic patterns.

Although tremendous efforts have been made to design sophisticated model architectures in order to fully capture the sophisticated spatial-temporal correlations, we still identify that almost all of the existing models are based on supervise learning, which usually require exhaustive handicraft labeling as prerequisite but have limited representation capability. The current benchmark supervised models ignore the reality of practical applications, thus significantly hindering their application scopes. Meanwhile, we notice that many renowned scholars have emphasized that the current limitations of deep learning are actually the limitations of supervise learning technology, and self-supervised learning is the future direction of Artificial Intelligence~\cite{12LeCun,STGCL,GCN-Survey}. 

Recently, self-supervised methods have shown great capability in representation learning tasks including natural language processing (NLP)~\cite{13dialogues}, image/video processing~\cite{14CV}, and recommendation system~\cite{15recommend}. The core idea of self-supervised learning is to derive some auxiliary supervised signals from the input dataset itself, which is capable of exploring the hidden distribution and patterns of the dataset. Among the various self-supervised methods, contrastive learning-based methods always demonstrated superior performance, such as MoCo~\cite{16MoCo} in object detection/image segmentation, BERT-CL~\cite{13dialogues} for retrieval-based dialogues, and GraphCL~\cite{18GraphCL} for graph data classification. While some studies have extended contrastive learning for graph-structure data~\cite{17Hassani,18GraphCL}, there are still three major problems worthy of further investigation. First, existing spatiotemporal contrastive learning models only adopt pre-defined static Graph Data Augmentation (GDA) approaches, which heavily limits the performance of the later contrastive representation learning. Second, previous research works~\cite{03STSGCN,06DGCRN} have proved the necessity and effectiveness of spatial-temporal dependency synchronous modeling. Nevertheless, none of the present graph contrastive models consider building simultaneous spatial-temporal dependency modeling structure for urban traffic forecasting. Third, we identify that none of the existing contrastive learning models realize both Graph-level and Node-level contrasting for a given spatiotemporal prediction task. 

To address the aforementioned problems, we propose a self-supervised model named Spatial-Temporal Synchronous Contextual Contrastive Learning (STS-CCL). The main contributions of this work are summarized as follows:
\begin{itemize}
\item We present a novel spatiotemporal contrastive learning model named STS-CCL for spatiotemporal graph representation learning and downstream prediction tasks. To begin with, we elaborately design two different yet correlated graph data augmentation methods to enhance contrastive learning performance. Furthermore, we present a hard mutual-view contrastive training scheme and formulated an integrated objective function to assist contrastive model training.
\item A spatial-temporal synchronous contrastive module (STS-CM) is proposed to realize spatial-temporal synchronous traffic dependency modeling and Graph-level contrasting, which addresses the spatial and temporal separate modeling caused inconsistency in current contrastive models. 
\item Aside from Graph-level contrasting, a comprehensive semantic contextual contrastive module (SC-CM) is proposed to achieve Node-level contrasting. SC-CM can further capture the hard distinguishable patterns along with negative filtering by using non-linear projection head and semantic contrastive loss.
\item We conduct extensive experiments and evaluations on two real-world urban traffic datasets from Hangzhou Metro System crowd-flow and Seattle freeway network traffic speed. Empirical study results demonstrate that STS-CCL can effectively capture the decent and dynamic spatiotemporal representations and consistently surpass other state-of-the-art methods. 

\end{itemize}

\section{Basic Definitions for Spatial-Temporal Traffic Forecasting}

\noindent \textbf{Definition 1: Traffic Network $g{\star}$.} Most of the urban traffic scenarios can be represented as weighted undirected graphs $g{\star}=(V{\star},E{\star},A{\star})$. In the graph representation, $V{\star}$ denotes a collection of vertices which containing all the node individuals in a given traffic network and $|V{\star}|=N$. $E{\star}$ denotes a collection of edges, representing the correlations between nodes. $A{\star}$ is the graph adjacency matrix of $g{\star}$ and $A{\star}=\{a_{ij}\}^{N \times N}$. Each matrix element $a_{ij}$ represents the computed closeness between node $v_i$ and $v_j$.

\noindent \textbf{Definition 2: Graph Adjacency Matrix $A{\star}$.} As the crucial component of a traffic graph representation, graph adjacency matrix has received lots of research attentions. In earlier studies, researchers usually adopt pre-defined static $A{\star}$, such as calculating the Euclidean distance or POI similarity between node $v_i, v_j$ as the matrix element $a_{ij}$. However, such time-invariant graph adjacency matrix fails to represent the dynamic change of a transportation network in real-world scenarios. Some later studies propose to use dynamic graph adjacency matrix and achieve superior performance~\cite{06DGCRN,10SCINet}.

\noindent \textbf{Definition 3: Spatiotemporal Traffic Forecasting based on Graph Structure Data.} With the historical traffic data sequence $X_{\tau-s+1:\tau}=[X_{\tau-s+1},X_{\tau-s+2},...,X_{\tau}]$ and the graph representation $g{\star}=(V{\star},E{\star},A{\star})$, spatiotemporal traffic forecasting task can be formulated as follows:
\begin{equation}
[X_{\tau-p+1:\tau}; g{\star}] \stackrel{f}{\rightarrow} \hat{X}_{\tau+1:\tau+k}
\label{eq:eq1}
\end{equation}

\noindent where $p$ denotes the total length of historical traffic series, $k$ denotes the prediction scale, $f$ is the learned mapping function of a neural network model.

\section{Methodology}

\begin{figure*}[htbp]
\centering
\includegraphics[width=0.75\textwidth]{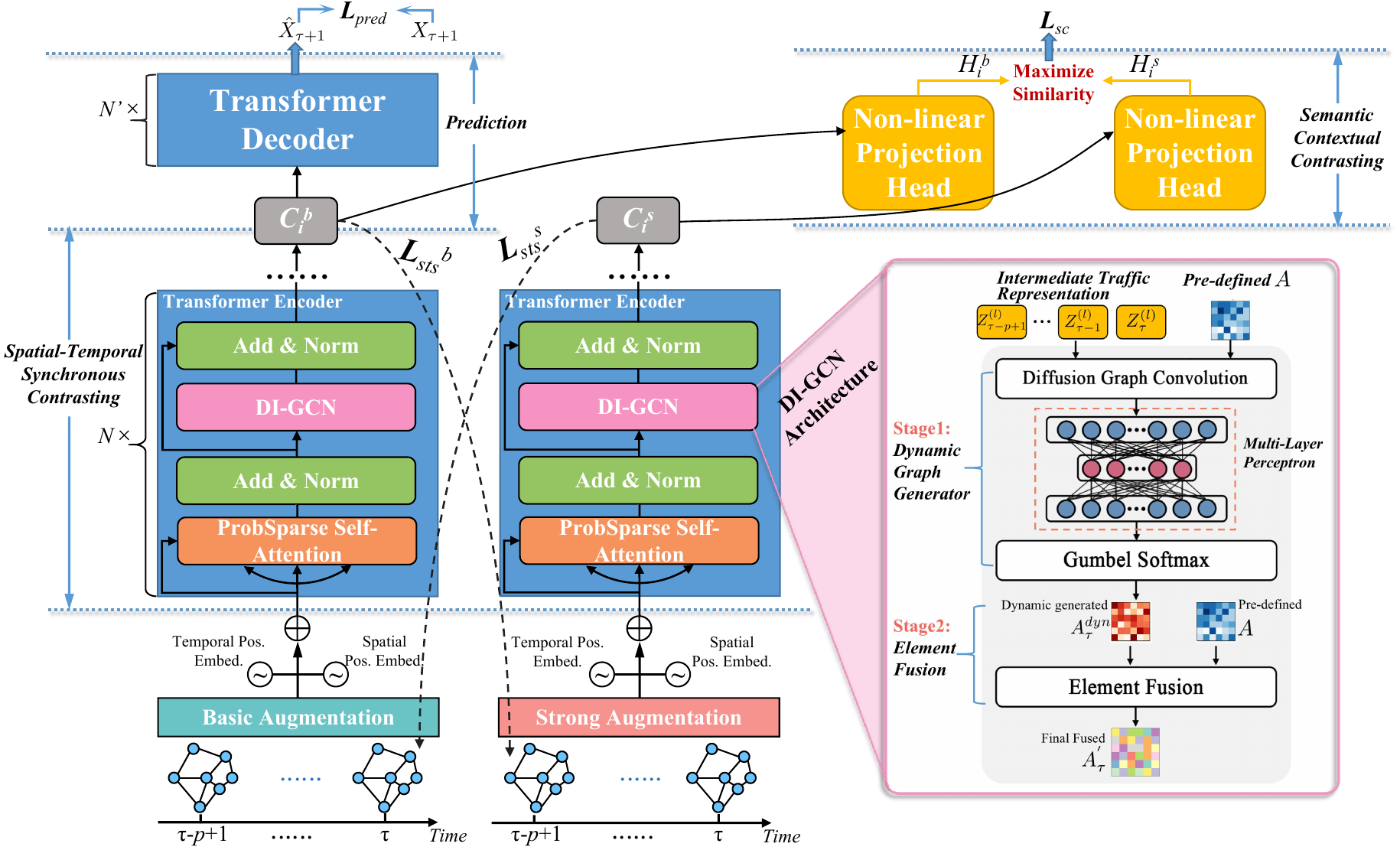}
\caption{Architecture of the proposed STS-CCL. We specifically illustrate the joint-learning scheme and the DI-GCN module.}
\label{fig1-model}
\end{figure*}
This section dedicates to introduce all the technical details of STS-CCL. As shown in Fig.~\ref{fig1-model}, STS-CCL adopts the joint-learning scheme, which consists of two branches (i.e., (1) traffic prediction branch, (2) contrastive learning branch), and the two branches are jointly learned during model training procedure. To begin with, we carry out data augmentation to generate the basic/strong augmentation views for the raw input traffic data. Next, the two views of data are separately fed into the STS-CM (denoted as "Transformer Encoder" in Fig.~\ref{fig1-model}) to achieve spatial-temporal synchronous contrastive learning with a designed hard mutual-view prediction task. Afterwards, the learned spatiotemporal traffic representations are used for two branch of tasks. The $C_t^{b}$ from basic augmentation view is sent to Transformer Decoder to generate future traffic predictions, the $C_t^{s}$ from strong augmentation view is sent to SC-CM to realize semantic contextual contrasting. Because of limited space, the temporal positional embedding and spatial positional embedding will be discussed in Appendix~\ref{appendA}. Finally, we introduce model training scheme and formulate the overall objective function.

\subsection{Spatiotemporal Graph Data Augmentation}
Data augmentation is an indispensable component and also the first step of contrastive learning models. Some latest works have investigated time-series data augmentation~\cite{TS-TCC} and graph data augmentation~\cite{18GraphCL} respectively, but there still lacks a professional spatiotemporal graph data augmentation methodology.

\begin{figure}[htbp]
\centering
\includegraphics[width=\columnwidth]{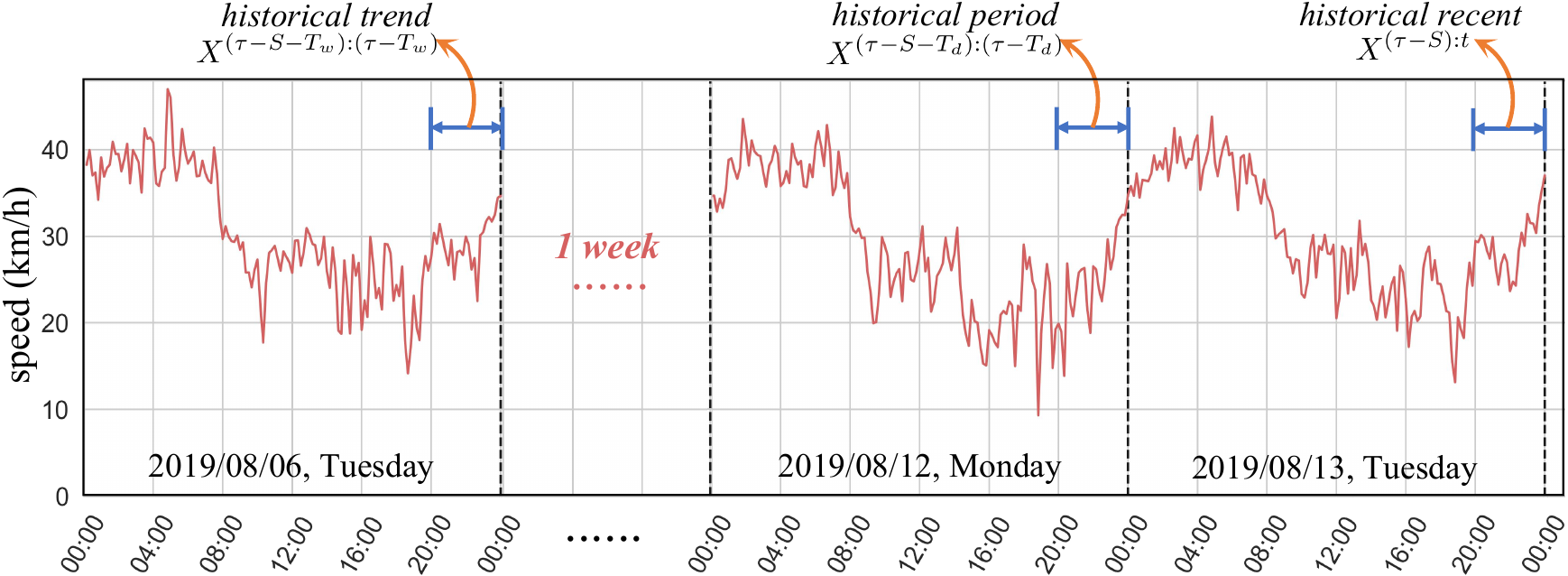}
\caption{Demonstration of the temporal scale fusion method.}
\label{fig2}
\end{figure}

In STS-CCL, we first adopt two latest developed graph data augmentations, namely Edge Perturbation (optimized version for spatiotemporal task) and Attribute Masking to augment data in terms of graph structure. As shown in Fig.~\ref{fig2}, we then propose a new \textbf{Temporal Scale Fusion method} to enhance data in terms of temporal characteristics. Additionally, we employ a learning-based graph view generator to adaptively create more underlying data views. By utilizing the above augmentation methods, we can create Basic and Strong Augmentation views considering both spatiotemporal characteristics and dynamic graph-data features. 

\subsubsection{Basic Augmentation.} For Basic Augmentation, we consider using (1)Edge masking, (2)Attribute masking, and (3)Temporal scale fusion as the augmentation techniques. Edge masking is an optimized version of edge perturbation for spatiotemporal tasks, which suggest discarding a specific ratio of edges to modify graph structures and is implemented by masking a specific ratio of adjacency matrix elements as 0~\cite{STGCL}. Attribute masking can also be referred from~\cite{STGCL}, which is directly implemented by randomly masking a specific ratio of input spatiotemporal data as 0.

Next, we introduce the proposed Temporal Scale Fusion method. The multi-scale temporal nature (i.e. historical recent/historical period/historical trend) of spatiotemporal traffic data has been proved in previous researches. However, it has been less explored in existing contrastive learning models. Effectively integrating traffic data from the three-granularity temporal scales can excavate more diverse spatiotemporal patterns. In considering that, we obtain the correlated data from historical recent/period/trend and fuse them together. FIg.~\ref{fig2} illustrates how the temporal scale fusion method operates. Let $X_{\tau-S:\tau}$ be the traffic data sequence from the historical recent time, $T_d$ and $T_w$ represent the time gap of historical period and trend, respectively. $X^{(\tau-S-T_d):(\tau-T_d)}$ means the traffic data comes from the same time window of the last day, $X^{(t-S-T_w):(t-T_w)}$ means the traffic data comes from the same time window of the last week. Temporal Scale Fusion method is formulated as in Eq.~\ref{eq:eq3}.

\begin{equation}
\small
\begin{aligned}
\tilde{X}^{(\tau-S):\tau}&=(1-\alpha-\beta)X^{(\tau-S):t}+\alpha X^{(\tau-S-T_d):(\tau-T_d)}\\
& +\beta X^{(\tau-S-T_w):(\tau-T_w)}
\end{aligned}
\label{eq:eq3}
\end{equation}

\noindent where $\tilde{X}^{(\tau-S):\tau}$ denotes the multi-scale temporal fused traffic data, $\alpha$ and $\beta$ are hyper-parameters. $\alpha,\beta$ are generated from the distribution $U(\delta_{ts},1)$ and then divided by two. $\delta_{ts}$ is a tunable factor, which ensures every epoch training data has their unique $\alpha$ and $\beta$.(i.e. this method is input-specific).

\subsubsection{Strong Augmentation.} 
\begin{figure}[htbp]
\centering
\includegraphics[width=\columnwidth]{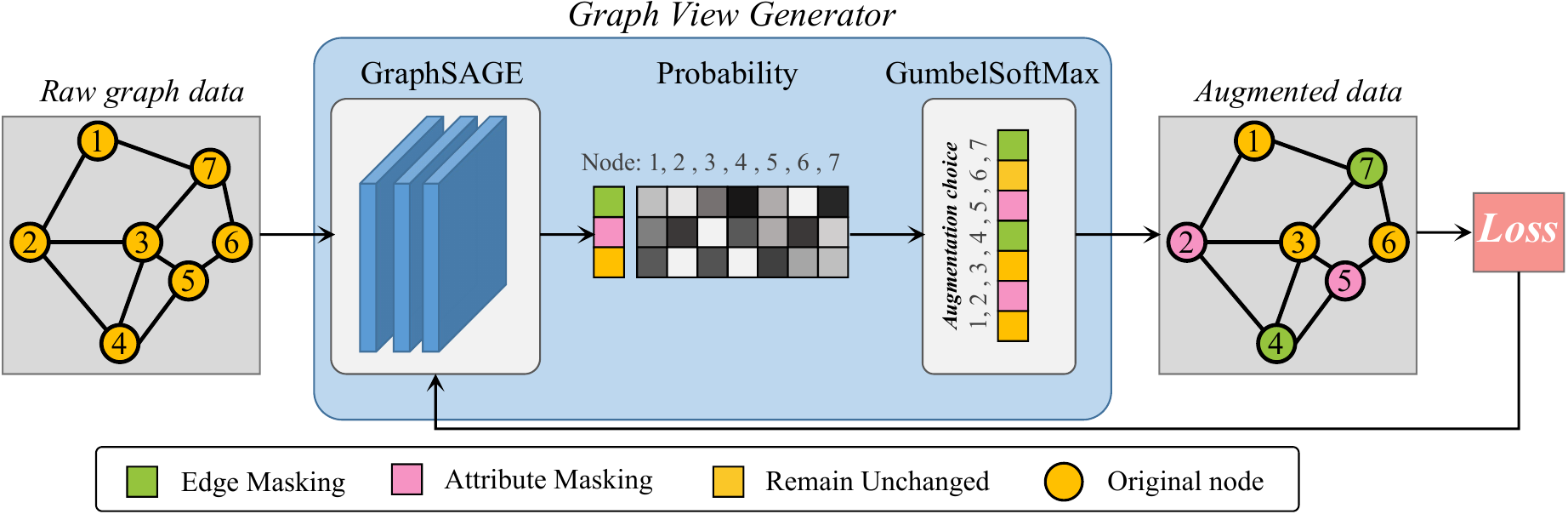}
\caption{The architecture of learning-based graph view generator.}
\label{fig3}
\end{figure}

We propose to use a learning-based graph view generator for dynamic and effective augmentation, as illustrated in Fig.~\ref{fig3}. For each node individual, the graph view generator first obtains the embedded node feature using GNN layers. Here, the GNN network is adopted following GraphSAGE~\cite{GraphSAGE}. The node embeddings are then transformed into probability distributions of choosing a specific graph data augmentation technique. There are three augmentation choices for each node: (1) Edge masking, (2) Attribute masking, and (3) Remain unchanged. Next, we use GumbelSoftMax upon the probability distributions to generate the final choice for each node individual. Finally, the selected best augmentation technique is applied on each node. Let $h_{i}^{(l)}$ and $e_{i}^{(l)}$ be the hidden state and node embedding of $i$-th node at the $l$-th GNN layer, respectively. $X_i$ denotes the node features and $F_i$ denotes the graph augmentation choice of node $i$. The learnable graph view generation procedure can be formulated as:
\begin{small}
\begin{align}
h_{i}^{(l-1)}&=COMBINE^{(l)}(h_i^{(l-2)},e_{i}^{(l-1)}) \\ 
e_{i}^{(l-1)}&=AGGREGATE^{(l)}([h_j^{(l-1)}: j \in N(i)]) \\ 
F_i&=GumbelSoftMax(e_{i}^{(l)}) \\ 
\tilde{X}_{i}&=Multiply(X_i,F_i) 
\end{align}
\end{small}

\noindent where Eq. 4 and Eq. 5 are the GNN node embedding process. $e_{i}^{(l-1)}$ denotes the probability distribution of choosing every augmentation techniques. $F_i$ is the one-hot vector generated by GumbelSoftMax, which indicates the final choice of augmentation. $Multiply(X_i,F_i)$ denotes the implemented node individual augmentation function. $X_i$ is the finally augmented node features. 

After the augmentation process from learnable graph view generator, the next step is to employ temporal scale fusion method for further processing, which finally result in the strong augmentation view data.

\subsection{Spatial-Temporal Synchronous Contrastive Learning}
This subsection dedicates to develop the proposed spatial-temporal synchronous contrastive module (STS-CM). STS-CM is constructed upon Transformer architecture, and we stack $N$ identical STS-CM in spatial-temporal synchronous contrasting stage. STS-CM is denoted as Transformer Encoder in Fig.~\ref{fig1-model}, which basically consists of two components: the ProbSparse self-attention mechanism~\cite{Informer} and the Dynamic Interaction GCN (DI-GCN). In each STS-CM module, the ProbSparse self-attention is capable of capturing the dynamic traffic dependencies among the temporal dimension, while our designed DI-GCN is responsible of capturing the dynamic interactive spatial dependencies among the spatial dimension. Therefore, the spatial-temporal synchronous traffic dependency modeling is achieved. Furthermore, we let the model to carry out a hard mutual-view forecasting objective in the synchronous contrastive learning stage after the $N \times$ STS-CM traffic feature modeling. The hard mutual-view forecasting is to use the extracted traffic feature from the Basic Augmentation data to predict the future one view of the Strong Augmentation, and vice versa. This training scheme can boost the contrastive model performance upon the spatial-temporal synchronous modeling. 

\subsubsection{ProbSparse self-attention.} The ProbSparse self-attention used in STS-CM can be referred from Informer~\cite{Informer}, which decreases the time and space complexity from $O(L^2)$ (vanilla self-attention) to $O(L \ln L)$, yielding efficiency and improved model performance.

\subsubsection{Dynamic Interaction GCN} The source idea of DI-GCN comes from the identification that most of existing graph adjacency matrix construction methods are pre-defined and time-invariant. We propose Dynamic Interaction Graph Convolutional Networks (DI-GCN) to adaptively adjust the graph adjacency matrix by dynamically interact with traffic data from different time intervals. The right part of Fig.~\ref{fig1-model} illustrates how DI-GCN adaptively generates the graph adjacency matrix $A_{\tau}^{'}$ (i.e. the "Final Fused $A_{\tau}^{'}$") at time step $\tau$. It can be seen that the generation of the final fused $A_{\tau}^{'}$ includes two major stages: (1) Dynamic Graph Generator, and (2) Element Fusion.

As shown in Fig.~\ref{fig1-model}, the Dynamic Graph Generator is composed of diffusion graph convolution (DGC) layer, multi-layer perceptron (MLP) and GumbelSoftmax. In the $l$-th Transformer encoder, the learned traffic representations after ProbSparse attention: $Z^{(l)}=(Z_{\tau-p+1}^{(l)},Z_{\tau-p+2}^{(l)},...,Z_{\tau}^{(l)})$ and the pre-defined graph adjacency matrix are fed into diffusion graph convolution to extract features, then fed into MLP and obtain an intermediate graph adjacency matrix $\tilde{A}_{\tau}^{dyn}$, which can be formulated as:
\begin{equation}
\small
\tilde{A}_{\tau}^{dyn}=Softmax(MLP(DGC(Z^{(l)},A)))
\label{eq:eq9}
\end{equation}
\noindent where MLP is the multi-layer perceptron, DGC is the diffusion graph convolution. 

Given that we need to sample $A_{\tau}^{dyn}$ during model training, but the intermediate graph adjacency matrix $\tilde{A}_{\tau}^{dyn}$ is actually discrete. To ensure the sampling process is derivable, we adopt GumbelSoftmax to re-parameterize the intermediate graph adjacency matrix, and obtain the final dynamic graph adjacency matrix $A_{\tau}^{dyn}$:
\begin{equation}
\small
\begin{aligned}
A_{\tau}^{dyn}&=GumbelSoftmax(\tilde{A}_{\tau}^{dyn}) \\
&=Softmax(\log(\tilde{A}_{\tau}^{dyn})-\log(-\log(\mu))/\Omega)
\end{aligned}
\label{eq:eq10}
\end{equation}
\noindent where $\mu \in Gumbel(0,1)$ is a random variable, $\Omega$ is a parameter to adjust the Softmax function, we set $\Omega=0.5$ in this work. The generated $A_{\tau}^{dyn}$ has strong capability to represent the dynamic changing correlations between nodes in a traffic network. 

Next, we integrate the generated dynamic $A_{\tau}^{dyn}$ with the pre-defined $A$ using element-wise dot product to obtain the final fused graph adjacency matrix $A_{\tau}^{'}$: $A_{\tau}^{'}=A \odot A_{\tau}^{dyn}$. Specifically, $A$ is the static connectivity adjacency matrix (i.e. $A={a_{ij}^{N \times N},a_{ij} \in \{0,1\}}$). $A_{\tau}^{'}$ will replace the original $A$ to participate in DI-GCN operation, as formulated below:
\begin{equation}
\small
\begin{aligned}
X_{t}^{(l)}&=\text{DI-GCN}(X_t^{(l-1)})=\sigma(A^{'}X_t^{(l-1)}W^{(l)})\\
&=\sigma((A \odot A_{\tau}^{dyn})X_t^{(l-1)}W^{(l)})
\end{aligned}
\label{eq:eq11}
\end{equation}
\noindent where $X_t^{(l-1)}$ denotes the input traffic representations of DI-GCN module, $W^{(l)}$ denotes weight parameters of DI-GCN module, $X_t^{(l)}$ denotes the output results of DI-GCN module in the $l$-th Transformer encoder.

\subsubsection{Hard mutual-view training scheme.}The spatial-temporal synchronous contrastive learning stage adopts a hard mutual-view prediction task, which employs the learned representations from one augmentation view to predict the future data of another augmentation view. Let $\bm{Z}$ be the input of Transformer encoder after data augmentation and positional embedding procedures. The Transformer encoder summarizes all the historical representations $Z_{\tau-p+1:\tau}$ to obtain a spatiotemporal synchronous contrastive vector $C_{\tau}, C_{\tau}=f_{ENC}(Z_{\tau-p+1:\tau})$. The spatiotemporal synchronous contrastive vector $C_{\tau}$ is then used to predict the future input representations from $Z_{\tau+1:\tau+k}$. In STS-CCL model, the basic augmentation view generates $C_{\tau}^b$ and the strong augmentation view generates $C_{\tau}^s$. Then, we design a hard mutual-view prediction task by employing the learned contrastive vector $C_{\tau}^b$ to predict the future input representations of the strong augmentation view: $Z_{\tau+1:\tau+k}$, and vice versa. The spatial-temporal synchronous contrastive loss $L_{sts}^{*}, * \in \{b,s\}$ here is defined by minimizing the dot product of a predicted representation and the corresponding ground-truth representation, while maximizing the dot product with other negative samples $N_{\tau,k}$. Thus, the spatial-temporal synchronous contrastive loss for the two views are formulated as follows:

\begin{align}
\small
L_{sts}^s&=-\frac{1}{K} \sum_{k=1}^{K} \log \frac{\exp ((W_k (C_{\tau}^s))^{\mathrm{T}}Z_{\tau+k}^b)}{\sum_{n \in N_{\tau,k}} \exp (((W_k (C_{\tau}^s))^{\mathrm{T}})Z_n^b)} \\ 
L_{sts}^b&=-\frac{1}{K} \sum_{k=1}^K \log \frac{\exp ((W_k (C_{\tau}^b))^{\mathrm{T}}Z_{\tau+k}^s)}{\sum_{n \in N_{\tau,k}} \exp((W_k(C_{\tau}^b))^{\mathrm{T}}Z_{n}^s)} 
\end{align}
\noindent where $W_k$ is a linear function used to reshape $C_{\tau}^{*}$'s dimension to be the same as $Z$, and the Log Bi-linear equation is adopted in the spatial-temporal synchronous contrastive loss formulation. 

The finally learned basic/strong view contrastive vector from the $N$-th Transformer encoder will be sent to execute traffic prediction and stage-two contrastive learning (semantic contextual contrasting), respectively. 

\subsection{Semantic Contextual Contrastive Learning}
As far as we are concerned, existing contrastive models simply treat all other samples aside from node $i$ itself as the negative samples to be contrasted. However, the nodes in a traffic network may have similar spatial heterogeneity or similar semantic contexts. In this part, we unify similar spatial heterogeneity and similar semantic contexts as positive samples and filter them out in the denominator of semantic contextual contrastive loss. Given a batch of $N$ nodes in model training, the learned representation by the non-linear projection head $H_{i}^b$ and it's corresponding representation from another view $H_{i}^{s}$, a general form of semantic contextual contrastive loss can be defined as Eq.~\ref{eq:eq14}:
\begin{equation}
L_{sc}=\frac{1}{N} =-\log \frac{\exp(sim(H_{i}^b,H_{i}^s)/\Delta)}{\sum_{j \in \mathbb{N}_{i}} \exp(sim(H_{i}^b,H_{j}^s)/\Delta)} 
\label{eq:eq14}
\end{equation}

\noindent where $sim(a,b)=\frac{a^T b}{|\!|a|\!| |\!|b|\!|}$ denotes the dot product between $l_2$ normalized $a$ and $b$ (i.e. cosine similarity), $\Delta$ is a controlled temperature parameter. $\mathbb{N}_{i}$ denotes the determined acceptable negative sample collection, which is detailed introduced in the following paragraphs. 

Specifically, for spatial heterogeneity similarity filtering, we use the well-known node connectivity adjacency matrix $A_{con}$~\cite{GCN-Survey} to measure the spatially similarity. $A_{con}=(a_{ij})^{N \times N} \in \{0,1\}$ is defined as follows: if node $i$ and $j$ (e.g. sensors, traffic stations, road segments, etc.) are geographically neighbors, the corresponding element $a_{ij}$ is set as 1, otherwise 0. Therefore, we obtain the similar spatial heterogeneity samples for each node within a batch, and exclude these data samples for negative calculation. 

In terms of semantic context similarity filtering, we first assign a semantic vector $M_{i,\tau}$ for each node, which consists of POI distribution information (including seven categories of POIs: school, hospital, commercial center, shopping mall, stadium, transportation centre, scenic spot) and refined timestamp information (including (a).day-of-week, (b).is-weekend, (c).is holiday). All these semantic features are encoded using One-Hot Encoding method. Next, we introduce how to calculate the semantic context similarity between nodes. For the first step, we calculate the similarity score between two nodes using Jensen-Shannon (JS) divergence~\cite{Zhou2020}, which can be formulated as:

\begin{eqnarray} \label{eq:eq15}
\small
    \begin{array}{lr}
    \text{Sim}(M_{i,\tau},M_{j,\tau})\!=\!1-\text{JS}(M_{i,\tau},M_{j,\tau}), & \\
    \text{JS}(M_{i,\tau},M_{j,\tau})\!=\! \frac{1}{2} \!\sum \limits_{1\leq q\leq Q} \bigg( ^{M_{i,\tau}(q)\log\frac{2M_{i,\tau}(q)}{M_{i,\tau}(q)+M_{j,\tau}(q)}+}_{M_{j,\tau}(q) \log\frac{2M_{j,\tau}(q)}{M_{i,\tau}(q)+M_{j,\tau}(q)}}\bigg)
    \end{array}
\end{eqnarray}

\noindent where $M_{i,\tau}, M_{j,\tau} \in \mathbb{R}^{Q}$ represents the semantic vector of node $i$ and $j$, of which sum equals to 1. $M_{i,\tau}(q)$ means the $q$-th dimension of semantic vector $M_{i,\tau}$. 

For the second step, we select the Top-u most semantically similar nodes and exclude the Top-u nodes from the negative samples. From the above spatial heterogeneity filtering and semantic context similar filtering, we can finally obtain the acceptable negative sample collection $\mathbb{N}_{i}$, which is applied to our semantic contextual contrastive loss as in Eq.~\ref{eq:eq14}.

\subsection{Loss Function and Model Training Method}
We employ joint-learning scheme for STS-CCL model, which means the model conduct prediction task and contrastive learning task in a simultaneous manner. The basic augmentation view is not only used for stage-one contrastive learning (i.e. spatial-temporal synchronous contrasting), but also used for the subsequent traffic prediction task. The strong augmentation view goes through stage-one contrastive learning together with basic augmentation view, then the two generated representations are fed into stage-two contrastive learning, which finalize the contrastive learning process. In our case, Transformer encoder $E_{\theta}(\cdot)$ is jointly trained with Transformer decoder $D_{\omega}(\cdot)$ and the non-linear projection head $g_{\phi}(\cdot)$ in semantic contextual contrasting. The contrastive loss can serve as additional self-supervised signals in the overall objective function and improve spatiotemporal traffic forecasting performance. 

Since STS-CCL adopts the joint-learning scheme, its overall objective function should consists of both traffic prediction loss $L_{pred}$ and contrastive learning loss $L_{cl}$. We determine \textbf{the overall objective function} as: $L_{\text{STS-CCL}}=L_{pred}+\epsilon L_{cl}$, where $\epsilon$ is a weight-parameter to adaptively balance the importance of the two parts. As shown in Fig.~\ref{fig1-model}, $L_{pred}$ is calculated between the generated future traffic prediction from the Transformer Decoder and the ground-truth future traffic data. We adopt the Mean Squared Error (MSE) as traffic prediction loss, which can be formulated as: $L_{pred}=\frac{1}{M}\sum_{i=1}^M (X_i-\hat{X}_i)^2$, where $M$ is the total number of testing data samples. The contrastive learning loss $L_{cl}$ includes spatial-temporal synchronous contrasting loss $L_{sts}^{b}, L_{sts}^{s}$ and semantic contextual contrasting loss $L_{sc}$, thus could be formulated as: $L_{cl}=L_{sts}^{b}+L_{sts}^{s}+L_{sc}$. The STS-CCL model training algorithm is provided in Appendix~\ref{appendC}. 

\section{Experimental Evaluations}
\noindent \textbf{Datasets.} We adopt two popular open-source datasets in spatial-temporal traffic data mining, namely Hangzhou-Metro\footnote{\url{https://tianchi.aliyun.com/competition/entrance/231708/information}} and Seattle-speed\footnote{\url{https://github.com/zhiyongc/Seattle-Loop-Data}}. Hangzhou-Metro is a crowd-flow mobility dataset collected from the metro system of Hangzhou city within a month in year 2019. The raw dataset contains over 70 million pieces of passenger records at seconds level, we process the raw data to represent the 81 subway stations' inflow/outflow in every 10-minute. Seattle-speed is a traffic network speed dataset collected by the distributed loop detectors on the freeways in Seattle area in year 2015. Table~\ref{tab1} summarizes the statistics of the two datasets. 
\begin{table}[!htb]
\footnotesize
  \caption{Statistics of the two datasets used in experiments.}
  \label{tab1}
\resizebox{\columnwidth}{10mm}{
  \begin{tabular}{lcc}
    \hline
    \textbf{Dataset} &\textbf{Hangzhou-Metro} &\textbf{Seattle-speed} \\
    \hline
    \tabincell{l}{Time range} & 01/01/2019-01/27/2019 & 01/01/2015-12/31/2015 \\
    \tabincell{l}{Total data sample} & 314,928 & 105,120 \\
    \tabincell{l}{Time interval} & 10-minute & 5-minute \\
    \tabincell{l}{Nodes number} & 81 & 323 \\
    \hline
  \end{tabular}}
\end{table}

\noindent \textbf{Baseline Methods.} The proposed STS-CCL is compared with: i) Recently proposed deep learning-based models that demonstrated top performance: DCRNN~\cite{08DCRNN}, STSGCN~\cite{03STSGCN}, ASTGNN~\cite{ASTGNN}, AGCRN~\cite{05AGCRN}, RGSL~\cite{RGSL}, STG-NCDE~\cite{11STG-NCDE}. ii) State-of-the-art contrastive learning model in traffic forecasting: STGCL~\cite{STGCL} and SPGCL~\cite{SPGCL}. iii) ST-GSP~\cite{07ST-GSP}: a self-supervised learning model for urban flow prediction. We adopt root mean squared error (RMSE), mean absolute error (MAE) and mean absolute percentage error (MAPE) as the \textbf{Evaluation Metrics}. The implementation details are provided in Appendix~\ref{appendB} because of limited space.


\subsection{Results Comparison}
\begin{table*}[htbp]
\centering
\footnotesize
\caption{Performance comparison among all the baseline methods and STS-CCL on Hangzhou-Metro and Seattle-speed datasets.}
\label{tab2}
\resizebox{\linewidth}{!}{
\begin{tabular}{cc|ccc|ccc}
\hline
\multicolumn{2}{c|}{\multirow{2}{*}{Methods}}                                     & \multicolumn{3}{c|}{Seattle-speed}                                            & \multicolumn{3}{c}{Hangzhou-Metro}                                                                                                   \\ \cline{3-8} 
\multicolumn{2}{c|}{}                                                                                                              & \multicolumn{1}{c|}{RMSE}                       & \multicolumn{1}{c|}{MAE}                        & MAPE                           & \multicolumn{1}{c|}{RMSE}                        & \multicolumn{1}{c|}{MAE}                         & MAPE                            \\ \hline
\multicolumn{1}{c|}{\multirow{6}{*}{\makecell[c]{Recently proposed\\ deep learning-based\\models}}}                       & DCRNN                   & \multicolumn{1}{c|}{4.95±0.43}                  & \multicolumn{1}{c|}{2.73±0.21}                  & 6.68\%±0.25\%                  & \multicolumn{1}{c|}{40.51±1.39}                  & \multicolumn{1}{c|}{23.48±0.96}                  & 15.82\%±0.75\%                  \\ 
\multicolumn{1}{c|}{}                                                                                    & STSGCN                 & \multicolumn{1}{c|}{5.12±0.48}                  & \multicolumn{1}{c|}{2.85±0.25}                  & 7.07\%±0.29\%                  & \multicolumn{1}{c|}{40.17±1.47}                  & \multicolumn{1}{c|}{23.95±0.87}                  & 16.33\%±0.58\%                  \\ 
\multicolumn{1}{c|}{}                                                                                    & ASTGNN                  & \multicolumn{1}{c|}{4.71±0.51}                  & \multicolumn{1}{c|}{2.69±0.22}                  & 6.53\%±0.17\%                  & \multicolumn{1}{c|}{39.64±1.15}                  & \multicolumn{1}{c|}{23.09±0.82}                  & 15.65\%±0.56\%                  \\ 
\multicolumn{1}{c|}{}                                                                                    & AGCRN                   & \multicolumn{1}{c|}{4.58±0.17}                  & \multicolumn{1}{c|}{2.82±0.10}                  & 6.17\%±0.15\%                  & \multicolumn{1}{c|}{37.86±0.61}                  & \multicolumn{1}{c|}{21.63±0.45}                  & 14.69\%±0.38\%                  \\ 
\multicolumn{1}{c|}{}                                                                                    & RGSL                    & \multicolumn{1}{c|}{5.06±0.23}                  & \multicolumn{1}{c|}{2.74±0.16}                  & 6.28\%±0.19\%                  & \multicolumn{1}{c|}{41.25±0.73}                  & \multicolumn{1}{c|}{23.31±0.56}                  & 15.87\%±0.43\%                  \\ 
\multicolumn{1}{c|}{}                                                                                    & STG-NCDE                & \multicolumn{1}{c|}{4.63±0.15}                  & \multicolumn{1}{c|}{2.65±0.09}                  & 6.35\%±0.15\%                  & \multicolumn{1}{c|}{39.42±0.48}                  & \multicolumn{1}{c|}{22.85±0.35}                  & 15.13\%±0.27\%                  \\ \hline
\multicolumn{1}{c|}{\multirow{2}{*}{\makecell[c]{Self-supervised\\learning model}}}                                     & \multirow{2}{*}{ST-GSP} & \multicolumn{1}{c|}{\multirow{2}{*}{5.08±0.22}} & \multicolumn{1}{c|}{\multirow{2}{*}{2.57±0.17}} & \multirow{2}{*}{6.51\%±0.22\%} & \multicolumn{1}{c|}{\multirow{2}{*}{39.53±0.55}} & \multicolumn{1}{c|}{\multirow{2}{*}{22.69±0.38}} & \multirow{2}{*}{15.30\%±0.34\%} \\
\multicolumn{1}{c|}{}                                                                                    &                         & \multicolumn{1}{c|}{}                           & \multicolumn{1}{c|}{}                           &                                & \multicolumn{1}{c|}{}                            & \multicolumn{1}{c|}{}                            &                                 \\ \hline
\multicolumn{1}{c|}{\multirow{2}{*}{\makecell[c]{State-of-the-art\\contrastive learning model}}} & SPGCL                   & \multicolumn{1}{c|}{4.53±0.16}                  & \multicolumn{1}{c|}{2.53±0.11}                  & 6.12\%±0.10\%                  & \multicolumn{1}{c|}{37.27±0.32}                  & \multicolumn{1}{c|}{20.84±0.27}                  & 14.46\%±0.23\%                  \\ 
\multicolumn{1}{c|}{}                                                                                    & STGCL                   & \multicolumn{1}{c|}{4.45±0.12}                  & \multicolumn{1}{c|}{2.71±0.07}                  & 6.25\%±0.13\%                  & \multicolumn{1}{c|}{36.69±0.34}                  & \multicolumn{1}{c|}{21.05±0.21}                  & 14.28\%±0.19\%                  \\ \hline
\multicolumn{1}{c|}{Our proposed model}                                                                  & STS-CCL                 & \multicolumn{1}{c|}{\textbf{4.22±0.07}}                  & \multicolumn{1}{c|}{\textbf{2.38±0.05}}                & \textbf{5.86\%±0.09\%}                  & \multicolumn{1}{c|}{\textbf{35.74±0.23}}                  & \multicolumn{1}{c|}{\textbf{20.37±0.17}}                & \textbf{13.96\%±0.08\%}                  \\ \hline
\multicolumn{2}{c|}{STS-CCL's improvement over the best baseline}                                                                  & \multicolumn{1}{c|}{+5.450\%}                   & \multicolumn{1}{c|}{+6.303\%}                   & +4.436\%                       & \multicolumn{1}{c|}{+2.658\%}                    & \multicolumn{1}{c|}{+3.338\%}                    & +2.292\%                        \\ \hline
\end{tabular}}
\end{table*}

Table~\ref{tab2} shows the crowd-flow/traffic speed forecasting results. We compare STS-CCL with the aforementioned three kinds of baselines. Not surprisingly, the deep learning-based models show inferior performance than contrastive learning-based models (i.e. STGCL, SPGCL, and STS-CCL), demonstrating the powerful representation learning ability of contrastive models. The self-supervised model ST-GSP outperforms several advanced deep learning-models (i.e. DCRNN, STSGCN, AGCRN, RGCL). STG-NCDE and ASTGNN achieve excellent performance among deep learning models. Compared within the contrastive learning models, STGCL proposes an optional node-level or graph-level contrastive learning, whereas STS-CCL employs a two-stage contrasting to achieve both node-level and graph-level contrastive learning. As for dynamic graph construction, SPGCL directly uses contrastive learning to build dynamic adjacency matrix for each time step, rather than using contrastive learning methods to capture the invariant traffic representations, which may loss the original intention of contrastive learning model. Furthermore, the model architecture of SPGCL is less elaborately designed, which leads to a trade-off performance. In addition, STS-CCL consistently shows superior performance over all the baseline methods.

\subsection{Ablation Study}
To evaluate the effectiveness of each model component as well as data augmentation, we proceed to compare STS-CCL with the following designed variants. Specifically, "STS-CM only" is trained without the hard mutual-view prediction task and the SC-CM module. "STS-CM+MVP" is trained adding the hard-mutual view prediction task but without the SC-CM module. The variant "STS-CM+MVP+SC-CM" denotes the proposed full STS-CCL model. "STS-CCL w/o negative filtering" is trained without using the proposed negative filtering method in SC-CM module, simply treating all other data samples as negatives. In "STS-CCL w/o DI-GCN", we replace the DI-GCN layer with a static GCN layer, which use the pre-defined connectivity adjacency matrix $A=(a_{ij})^{N \times N}, a_{ij} \in \{0,1\}$. Furthermore, we study the effectiveness of the proposed basic/strong data augmentation method. In "STS-CCL BA-only", the two different views of data are both generated using basic augmentation method, whereas in "STS-CCL SA-only", we generate the two views of data only using the strong augmentation method. Finally, we presents the ablation study results in Table~\ref{tab3}. It shows that without any key component in STS-CCL, the RMSE/MAE/MAPE results keep increasing, which suggest the degradation of model performance. Interestingly, we find that a specific model component may have different degree of impacts for different tasks, which can also be referred from Table~\ref{tab3}.

\begin{table*}[htbp]
\centering
\footnotesize
\caption{Evaluating the effectiveness of each model component and the data augmentation method adopted in STS-CCL.}
\label{tab3}
\resizebox{0.85\linewidth}{!}{
\begin{tabular}{c|ccc|ccc}
\hline
\multirow{2}{*}{Model Variants}                                           & \multicolumn{3}{c|}{Seattle-speed}                                                                   & \multicolumn{3}{c}{Hangzhou-Metro}                                                                     \\ \cline{2-7} 
                                                                          & \multicolumn{1}{c|}{RMSE}      & \multicolumn{1}{c|}{MAE}       & MAPE                               & \multicolumn{1}{c|}{RMSE}       & \multicolumn{1}{c|}{MAE}        & MAPE                                \\ \hline
STS-CM only                                                               & \multicolumn{1}{l|}{4.60±0.16} & \multicolumn{1}{l|}{2.56±0.13} & \multicolumn{1}{l|}{6.31\%±0.11\%} & \multicolumn{1}{l|}{38.15±0.42} & \multicolumn{1}{l|}{22.53±0.25} & \multicolumn{1}{l}{15.37\%±0.20\%} \\ 
STS-CM+MVP                                                                & \multicolumn{1}{l|}{4.48±0.13} & \multicolumn{1}{l|}{2.46±0.09} & \multicolumn{1}{l|}{6.05\%±0.13\%} & \multicolumn{1}{l|}{37.67±0.39} & \multicolumn{1}{l|}{21.45±0.24} & \multicolumn{1}{l}{14.58\%±0.15\%} \\ 
\begin{tabular}[c]{@{}c@{}}(\textbf{full STS-CCL}) \\ STS-CM+MVP+SC-CM\end{tabular} & \multicolumn{1}{c|}{\textbf{4.22±0.07}} & \multicolumn{1}{c|}{\textbf{2.38±0.05}} & \textbf{5.86\%±0.09\%}      & \multicolumn{1}{c|}{\textbf{35.74±0.23}} & \multicolumn{1}{c|}{\textbf{20.37±0.17}} & \textbf{13.96\%±0.08\%}               \\ 
\begin{tabular}[c]{@{}c@{}}STS-CCL w/o \\ negative filtering\end{tabular} & \multicolumn{1}{c|}{4.37±0.06} & \multicolumn{1}{c|}{2.41±0.07} & 5.93\%±0.12\%                      & \multicolumn{1}{c|}{36.08±0.25} & \multicolumn{1}{c|}{20.52±0.19} & 14.04\%±0.08\%                      \\ 
STS-CCL w/o DI-GCN                                                        & \multicolumn{1}{c|}{4.43±0.09} & \multicolumn{1}{c|}{2.52±0.06} & 6.08\%±0.11\%                      & \multicolumn{1}{c|}{36.85±0.27} & \multicolumn{1}{c|}{21.18±0.20} & 14.26\%±0.13\%                      \\ \hline
STS-CCL BA-only                                                           & \multicolumn{1}{c|}{4.56±0.10} & \multicolumn{1}{c|}{2.48±0.07} & 6.15\%±0.13\%                      & \multicolumn{1}{c|}{37.81±0.36} & \multicolumn{1}{c|}{21.33±0.19} & 14.35\%±0.12\%                      \\
STS-CCL SA-only                                                           & \multicolumn{1}{c|}{4.31±0.06} & \multicolumn{1}{c|}{2.43±0.05} & 5.96\%±0.10\%                      & \multicolumn{1}{c|}{36.52±0.28} & \multicolumn{1}{c|}{20.94±0.18} & 14.07\%±0.09\%                      \\ \hline
\end{tabular}
}
\end{table*}

\subsection{Robustness Study and Transferability}

\begin{figure}[htbp]
\centering
\includegraphics[width=0.99\columnwidth]{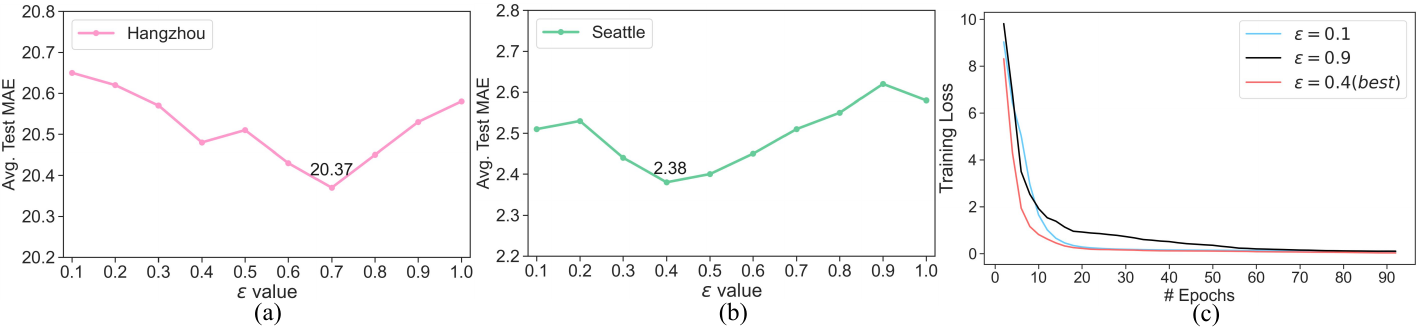}
\caption{(a)-(c):sensitivity study of $\epsilon$ in objective function.}
\label{fig5}
\end{figure}

\noindent \textbf{Sensitivity to $\epsilon$ in Objective Function.} We tune $\epsilon$ in the overall objective function of STS-CCL (see section 3.4) to evaluate the model robustness. Fig.~\ref{fig5}(a) and \ref{fig5}(b) visualize the MAE results under different $\epsilon$ settings. It can be easily seen that the optimal $\epsilon$ value is 0.70 and 0.40 for Hangzhou-Metro and Seattle-speed, respectively. Furthermore, Fig.~\ref{fig5}(c) shows the model training curve under two randomly selected $\epsilon$ value and the optimal value $\epsilon=0.40$ on Seattle-speed dataset. Regardless of the changing $\epsilon$ values, STS-CCL always achieve convergence within 40 epochs then become stable in the following epochs. Also, selecting the best $\epsilon$ can obviously assist model training process. 

\begin{figure}[htbp]
\centering
\includegraphics[width=0.75\columnwidth]{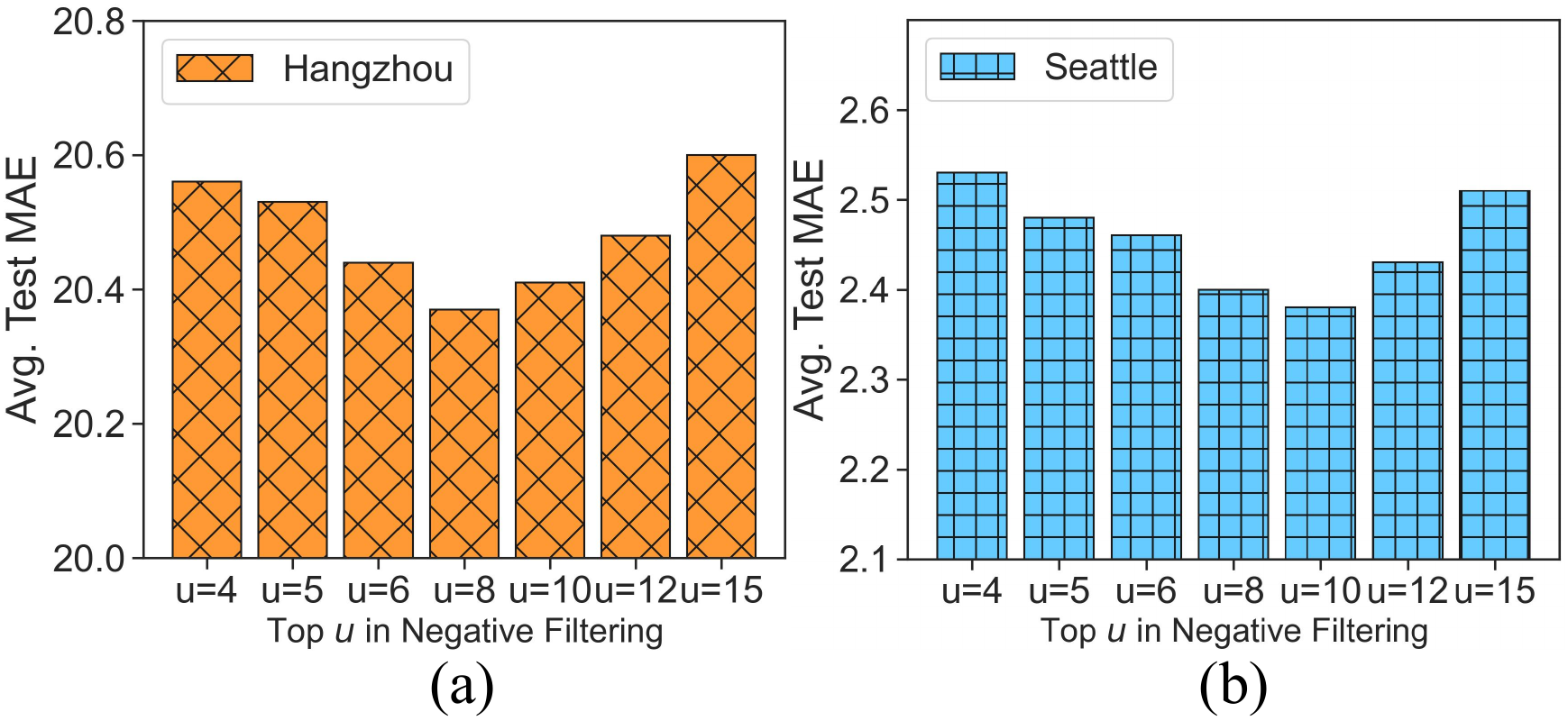}
\caption{(a)-(b): influence of Top-$u$ in Negative Filtering.}
\label{fig6}
\end{figure}
\noindent \textbf{Investigating $u$ and $A_{con}$ in Negative Filtering.} In this part, we first study how to set the Top-$u$ in semantic context similarity filtering. As shown in Fig.~\ref{fig6}(a)-Fig.~\ref{fig6}(b), we set $u=4,5,6,8,10,12,15$ and visualize the MAE performance under different Top-$u$. Second, the adopted $A_{con}$ in spatial heterogeneity similarity filtering is compared with another widely used distance-based adjacency matrix $A_{dist}$~\cite{08DCRNN}. We employ $A_{con}$ and $A_{dist}$ to carry out spatial heterogeneity similarity filtering respectively, and present the model performance comparison in Table~\ref{tab4}.

\begin{table}[htbp]
\centering
\scriptsize
\caption{Evaluating the efficiency of $A_{con}$ in Negative Filtering.}
\label{tab4}
\resizebox{\columnwidth}{8mm}{
\begin{tabular}{c|ccc|ccc}
\hline
Dataset                         & \multicolumn{3}{c|}{$A_{con}$}                                                                                                    & \multicolumn{3}{c}{$A_{dist}$}                                                                                                   \\ \hline
\multirow{2}{*}{Hangzhou-Metro} & \multicolumn{1}{c|}{RMSE}                       & \multicolumn{1}{c|}{MAE}                        & MAPE                           & \multicolumn{1}{c|}{RMSE}                       & \multicolumn{1}{c|}{MAE}                        & MAPE                           \\ \cline{2-7} 
                                & \multicolumn{1}{c|}{35.74±0.23}                 & \multicolumn{1}{c|}{20.37±0.17}                 & 13.96\%±0.08\%                 & \multicolumn{1}{c|}{36.13±0.22}                 & \multicolumn{1}{c|}{20.49±0.18}                 & 14.07\%±0.08\%                 \\ \hline
\multirow{2}{*}{Seattle-speed}  & \multicolumn{1}{c|}{\multirow{2}{*}{4.22±0.07}} & \multicolumn{1}{c|}{\multirow{2}{*}{2.38±0.05}} & \multirow{2}{*}{5.86\%±0.09\%} & \multicolumn{1}{c|}{\multirow{2}{*}{4.31±0.08}} & \multicolumn{1}{c|}{\multirow{2}{*}{2.44±0.05}} & \multirow{2}{*}{5.92\%±0.10\%} \\
                                & \multicolumn{1}{c|}{}                           & \multicolumn{1}{c|}{}                           &                                & \multicolumn{1}{c|}{}                           & \multicolumn{1}{c|}{}                           &                                \\ \hline
\end{tabular}
}
\end{table}

\begin{figure}[thbp]
\centering
\includegraphics[width=0.96\columnwidth]{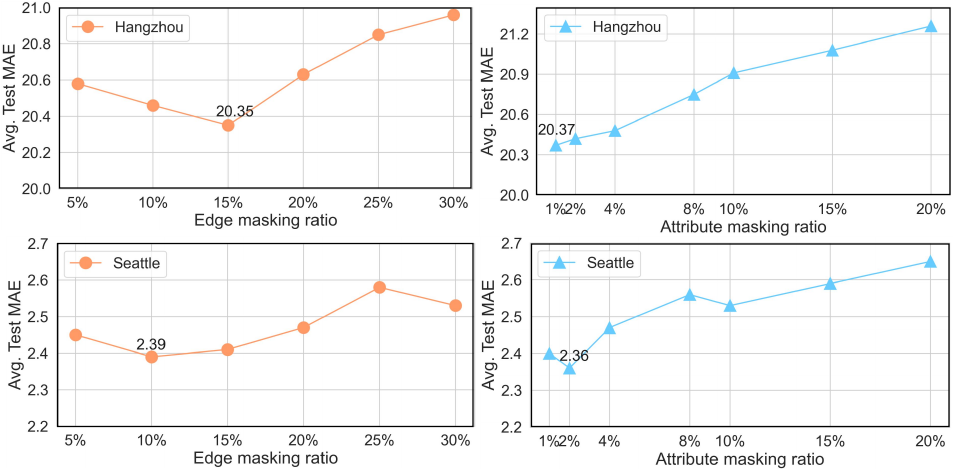}
\caption{Effects of different Edge/Attribute masking rate used in data augmentation.}
\label{fig4}
\end{figure}

\noindent \textbf{Effects of Edge/Attribute Masking Rate in Graph Data Augmentation.}
Here, we carry out spatiotemporal traffic prediction under different Edge/Attribute masking rates. The performance results are illustrated in Fig.~\ref{fig4}(a)-Fig.~\ref{fig4}(d), with the best Edge/Attribute masking rates been highlighted. We yield the following observations: First, regardless of the Edge/Attribute masking rates, our data augmentation methods always yield better performance than the SOTA contrastive learning traffic prediction methods. Second, although different datasets have their unique semantics and traffic features, the best-performed Edge/Attribute masking rates are quite stable, which verifies the transferability of STS-CCL model on different spatiotemporal datasets. Third, we find that Attribute masking has more significant influence on model performance, the reason is that masking a certain proportion of input data as 0 is similar to the traffic data missing scenario, which will directly impacts model training as well as the degradation of model performance.

\section{Conclusions}
This paper presents a general contrastive learning-based spatiotemporal forecasting model called STS-CCL, which for the first time realizes both node-level and graph-level contrasting, demonstrating strong representation learning ability over existing state-of-the-art methods. We first propose two general and efficient data augmentation methods for spatiotemporal contrastive learning. Then the designed STS-CM module simultaneously captures the decent spatial-temporal dependencies while achieving robust contrastive learning. We further propose a novel semantic contextual contrastive method for node-level contrasting, which overcomes the shortcoming of standard negative filtering in existing contrastive models. Extensive evaluations on two large-scale urban traffic datasets prove the superiority and effectiveness of STS-CCL over spatiotemporal benchmarks. We also provide comprehensive analysis and discussions for experimental results.

\begin{acks}
We thank Dr. Wei Dong for his valuable insights and discussions about the next generation's autonomous driving data mining during Lincan Li's tenure as a Senior R\&D Algorithm Engineer at BYD ghy Shenzhen, Dept. of IDA.
\end{acks}

\bibliographystyle{ACM-Reference-Format}
\bibliography{acm-mm-sigconf}


\begin{thebibliography}{28}


\ifx \showCODEN    \undefined \def \showCODEN     #1{\unskip}     \fi
\ifx \showDOI      \undefined \def \showDOI       #1{#1}\fi
\ifx \showISBNx    \undefined \def \showISBNx     #1{\unskip}     \fi
\ifx \showISBNxiii \undefined \def \showISBNxiii  #1{\unskip}     \fi
\ifx \showISSN     \undefined \def \showISSN      #1{\unskip}     \fi
\ifx \showLCCN     \undefined \def \showLCCN      #1{\unskip}     \fi
\ifx \shownote     \undefined \def \shownote      #1{#1}          \fi
\ifx \showarticletitle \undefined \def \showarticletitle #1{#1}   \fi
\ifx \showURL      \undefined \def \showURL       {\relax}        \fi
\providecommand\bibfield[2]{#2}
\providecommand\bibinfo[2]{#2}
\providecommand\natexlab[1]{#1}
\providecommand\showeprint[2][]{arXiv:#2}

\bibitem[12L(2020)]%
        {12LeCun}
 \bibinfo{year}{2020}\natexlab{}.
\newblock \bibinfo{title}{Talk Title: Self-Supervised Learning}.
\newblock
  \bibinfo{howpublished}{\url{https://aaai.org/Conferences/AAAI-20/invited-speakers/}}.
\newblock


\bibitem[Bai et~al\mbox{.}(2020)]%
        {05AGCRN}
\bibfield{author}{\bibinfo{person}{Lei Bai}, \bibinfo{person}{Lina Yao},
  \bibinfo{person}{Can Li}, \bibinfo{person}{Xianzhi Wang}, {and}
  \bibinfo{person}{Can Wang}.} \bibinfo{year}{2020}\natexlab{}.
\newblock \showarticletitle{Adaptive Graph Convolutional Recurrent Network for
  Traffic Forecasting}. In \bibinfo{booktitle}{\emph{Proceedings of the 34th
  International Conference on Neural Information Processing Systems}}
  \emph{(\bibinfo{series}{NIPS'20})}.
\newblock


\bibitem[Chen et~al\mbox{.}(2020)]%
        {14CV}
\bibfield{author}{\bibinfo{person}{Ting Chen}, \bibinfo{person}{Simon
  Kornblith}, \bibinfo{person}{Mohammad Norouzi}, {and}
  \bibinfo{person}{Geoffrey Hinton}.} \bibinfo{year}{2020}\natexlab{}.
\newblock \showarticletitle{A simple framework for contrastive learning of
  visual representations}. In \bibinfo{booktitle}{\emph{International
  conference on machine learning}}. \bibinfo{pages}{1597--1607}.
\newblock


\bibitem[Choi et~al\mbox{.}(2022)]%
        {11STG-NCDE}
\bibfield{author}{\bibinfo{person}{Jeongwhan Choi}, \bibinfo{person}{Hwangyong
  Choi}, \bibinfo{person}{Jeehyun Hwang}, {and} \bibinfo{person}{Noseong
  Park}.} \bibinfo{year}{2022}\natexlab{}.
\newblock \showarticletitle{Graph neural controlled differential equations for
  traffic forecasting}. In \bibinfo{booktitle}{\emph{Proceedings of the AAAI
  Conference on Artificial Intelligence}}, Vol.~\bibinfo{volume}{36}.
  \bibinfo{pages}{6367--6374}.
\newblock


\bibitem[Eldele et~al\mbox{.}(2021)]%
        {TS-TCC}
\bibfield{author}{\bibinfo{person}{Emadeldeen Eldele}, \bibinfo{person}{Mohamed
  Ragab}, \bibinfo{person}{Zhenghua Chen}, \bibinfo{person}{Min Wu},
  \bibinfo{person}{Chee Keong}, \bibinfo{person}{Xiaoli~Li Kwoh}, {and}
  \bibinfo{person}{Cuntai Guan}.} \bibinfo{year}{2021}\natexlab{}.
\newblock \showarticletitle{Time-Series Representation Learning via Temporal
  and Contextual Contrasting}. In \bibinfo{booktitle}{\emph{Proceedings of the
  30th International Joint Conference on Artificial Intelligence, {IJCAI-21}}}.
\newblock


\bibitem[Fang et~al\mbox{.}(2022)]%
        {01STAN}
\bibfield{author}{\bibinfo{person}{Ziquan Fang}, \bibinfo{person}{Dongen Wu},
  \bibinfo{person}{Lu Pan}, \bibinfo{person}{Lu Chen}, {and}
  \bibinfo{person}{Yunjun Gao}.} \bibinfo{year}{2022}\natexlab{}.
\newblock \showarticletitle{When Transfer Learning Meets Cross-City Urban Flow
  Prediction: Spatio-Temporal Adaptation Matters}. In
  \bibinfo{booktitle}{\emph{Proceedings of the Thirty-First International Joint
  Conference on Artificial Intelligence, {IJCAI-22}}}.
  \bibinfo{pages}{2030--2036}.
\newblock


\bibitem[Guo et~al\mbox{.}(2019)]%
        {09ASTGCN}
\bibfield{author}{\bibinfo{person}{Shengnan Guo}, \bibinfo{person}{Youfang
  Lin}, \bibinfo{person}{Ning Feng}, \bibinfo{person}{Chao Song}, {and}
  \bibinfo{person}{Huaiyu Wan}.} \bibinfo{year}{2019}\natexlab{}.
\newblock \showarticletitle{Attention based spatial-temporal graph
  convolutional networks for traffic flow forecasting}. In
  \bibinfo{booktitle}{\emph{Proceedings of the AAAI conference on artificial
  intelligence}}, Vol.~\bibinfo{volume}{33}. \bibinfo{pages}{922--929}.
\newblock


\bibitem[Guo et~al\mbox{.}(2022)]%
        {ASTGNN}
\bibfield{author}{\bibinfo{person}{Shengnan Guo}, \bibinfo{person}{Youfang
  Lin}, \bibinfo{person}{Huaiyu Wan}, \bibinfo{person}{Xiucheng Li}, {and}
  \bibinfo{person}{Gao Cong}.} \bibinfo{year}{2022}\natexlab{}.
\newblock \showarticletitle{Learning Dynamics and Heterogeneity of
  Spatial-Temporal Graph Data for Traffic Forecasting}.
\newblock \bibinfo{journal}{\emph{IEEE Transactions on Knowledge and Data
  Engineering}} \bibinfo{volume}{34}, \bibinfo{number}{11}
  (\bibinfo{year}{2022}), \bibinfo{pages}{5415--5428}.
\newblock


\bibitem[Hamilton et~al\mbox{.}(2017)]%
        {GraphSAGE}
\bibfield{author}{\bibinfo{person}{William~L. Hamilton},
  \bibinfo{person}{Zhitao Ying}, {and} \bibinfo{person}{Jure Leskovec}.}
  \bibinfo{year}{2017}\natexlab{}.
\newblock \showarticletitle{Inductive Representation Learning on Large Graphs}.
  In \bibinfo{booktitle}{\emph{Proceedings of the 31st International Conference
  on Neural Information Processing Systems}}
  \emph{(\bibinfo{series}{NIPS'17})}.
\newblock


\bibitem[Hassani and Khasahmadi(2020)]%
        {17Hassani}
\bibfield{author}{\bibinfo{person}{Kaveh Hassani} {and}
  \bibinfo{person}{Amir~Hosein Khasahmadi}.} \bibinfo{year}{2020}\natexlab{}.
\newblock \showarticletitle{Contrastive multi-view representation learning on
  graphs}. In \bibinfo{booktitle}{\emph{International Conference on Machine
  Learning}}. \bibinfo{pages}{4116--4126}.
\newblock


\bibitem[He et~al\mbox{.}(2020)]%
        {16MoCo}
\bibfield{author}{\bibinfo{person}{Kaiming He}, \bibinfo{person}{Haoqi Fan},
  \bibinfo{person}{Yuxin Wu}, \bibinfo{person}{Saining Xie}, {and}
  \bibinfo{person}{Ross Girshick}.} \bibinfo{year}{2020}\natexlab{}.
\newblock \showarticletitle{Momentum contrast for unsupervised visual
  representation learning}. In \bibinfo{booktitle}{\emph{Proceedings of the
  IEEE/CVF conference on computer vision and pattern recognition}}.
  \bibinfo{pages}{9729--9738}.
\newblock


\bibitem[Li et~al\mbox{.}(2022a)]%
        {06DGCRN}
\bibfield{author}{\bibinfo{person}{Fuxian Li}, \bibinfo{person}{Jie Feng},
  \bibinfo{person}{Huan Yan}, \bibinfo{person}{Guangyin Jin},
  \bibinfo{person}{Fan Yang}, \bibinfo{person}{Funing Sun},
  \bibinfo{person}{Depeng Jin}, {and} \bibinfo{person}{Yong Li}.}
  \bibinfo{year}{2022}\natexlab{a}.
\newblock \showarticletitle{Dynamic Graph Convolutional Recurrent Network for
  Traffic Prediction: Benchmark and Solution}.
\newblock \bibinfo{journal}{\emph{ACM Transactions on Knowledge Discovery from
  Data}} (\bibinfo{year}{2022}).
\newblock


\bibitem[Li et~al\mbox{.}(2018a)]%
        {Li2018deeper}
\bibfield{author}{\bibinfo{person}{Qimai Li}, \bibinfo{person}{Zhichao Han},
  {and} \bibinfo{person}{Xiao-Ming Wu}.} \bibinfo{year}{2018}\natexlab{a}.
\newblock \showarticletitle{Deeper insights into graph convolutional networks
  for semi-supervised learning}. In \bibinfo{booktitle}{\emph{Proceedings of
  the AAAI Conference on Artificial Intelligence}}.
\newblock


\bibitem[Li et~al\mbox{.}(2022b)]%
        {SPGCL}
\bibfield{author}{\bibinfo{person}{Rongfan Li}, \bibinfo{person}{Ting Zhong},
  \bibinfo{person}{Xinke Jiang}, \bibinfo{person}{Goce Trajcevski},
  \bibinfo{person}{Jin Wu}, {and} \bibinfo{person}{Fan Zhou}.}
  \bibinfo{year}{2022}\natexlab{b}.
\newblock \showarticletitle{Mining Spatio-Temporal Relations via Self-Paced
  Graph Contrastive Learning}. In \bibinfo{booktitle}{\emph{Proceedings of the
  28th ACM SIGKDD Conference on Knowledge Discovery and Data Mining}}.
  \bibinfo{pages}{936--944}.
\newblock


\bibitem[Li et~al\mbox{.}(2021)]%
        {Fourier-PE}
\bibfield{author}{\bibinfo{person}{Yang Li}, \bibinfo{person}{Si Si},
  \bibinfo{person}{Gang Li}, \bibinfo{person}{Cho-Jui Hsieh}, {and}
  \bibinfo{person}{Samy Bengio}.} \bibinfo{year}{2021}\natexlab{}.
\newblock \showarticletitle{Learnable Fourier Features for Multi-dimensional
  Spatial Positional Encoding}. In \bibinfo{booktitle}{\emph{Proceedings of the
  35th International Conference on Neural Information Processing Systems}}
  \emph{(\bibinfo{series}{NIPS'21})}.
\newblock


\bibitem[Li et~al\mbox{.}(2018b)]%
        {08DCRNN}
\bibfield{author}{\bibinfo{person}{Yaguang Li}, \bibinfo{person}{Rose Yu},
  \bibinfo{person}{Cyrus Shahabi}, {and} \bibinfo{person}{Yan Liu}.}
  \bibinfo{year}{2018}\natexlab{b}.
\newblock \showarticletitle{Diffusion Convolutional Recurrent Neural Network:
  Data-Driven Traffic Forecasting}. In \bibinfo{booktitle}{\emph{International
  Conference on Learning Representations (ICLR '18)}}.
\newblock


\bibitem[Lin et~al\mbox{.}(2019)]%
        {04DeepSTN+}
\bibfield{author}{\bibinfo{person}{Ziqian Lin}, \bibinfo{person}{Jie Feng},
  \bibinfo{person}{Ziyang Lu}, \bibinfo{person}{Yong Li}, {and}
  \bibinfo{person}{Depeng Jin}.} \bibinfo{year}{2019}\natexlab{}.
\newblock \showarticletitle{DeepSTN+: Context-Aware Spatial-Temporal Neural
  Network for Crowd Flow Prediction in Metropolis}. In
  \bibinfo{booktitle}{\emph{Proceedings of the AAAI Conference on Artificial
  Intelligence}}. \bibinfo{pages}{1020--1027}.
\newblock


\bibitem[Liu et~al\mbox{.}(2021)]%
        {10SCINet}
\bibfield{author}{\bibinfo{person}{Minhao Liu}, \bibinfo{person}{Ailing Zeng},
  \bibinfo{person}{Zhijian Xu}, \bibinfo{person}{Qiuxia Lai}, {and}
  \bibinfo{person}{Qiang Xu}.} \bibinfo{year}{2021}\natexlab{}.
\newblock \showarticletitle{Time series is a special sequence: Forecasting with
  sample convolution and interaction}.
\newblock \bibinfo{journal}{\emph{arXiv preprint arXiv:2106.09305}}
  (\bibinfo{year}{2021}).
\newblock


\bibitem[Liu et~al\mbox{.}(2022)]%
        {STGCL}
\bibfield{author}{\bibinfo{person}{Xu Liu}, \bibinfo{person}{Yuxuan Liang},
  \bibinfo{person}{Chao Huang}, \bibinfo{person}{Yu Zheng},
  \bibinfo{person}{Bryan Hooi}, {and} \bibinfo{person}{Roger Zimmermann}.}
  \bibinfo{year}{2022}\natexlab{}.
\newblock \showarticletitle{When Do Contrastive Learning Signals Help
  Spatio-Temporal Graph Forecasting?}. In \bibinfo{booktitle}{\emph{Proceedings
  of the 30th International Conference on Advances in Geographic Information
  Systems}} \emph{(\bibinfo{series}{SIGSPATIAL '22})}.
\newblock


\bibitem[Song et~al\mbox{.}(2020)]%
        {03STSGCN}
\bibfield{author}{\bibinfo{person}{Chao Song}, \bibinfo{person}{Youfang Lin},
  \bibinfo{person}{Shengnan Guo}, {and} \bibinfo{person}{Huaiyu Wan}.}
  \bibinfo{year}{2020}\natexlab{}.
\newblock \showarticletitle{Spatial-Temporal Synchronous Graph Convolutional
  Networks: A New Framework for Spatial-Temporal Network Data Forecasting}. In
  \bibinfo{booktitle}{\emph{Proceedings of the AAAI Conference on Artificial
  Intelligence}}. \bibinfo{pages}{914--921}.
\newblock


\bibitem[Xu et~al\mbox{.}(2021)]%
        {13dialogues}
\bibfield{author}{\bibinfo{person}{Ruijian Xu}, \bibinfo{person}{Chongyang
  Tao}, \bibinfo{person}{Daxin Jiang}, \bibinfo{person}{Xueliang Zhao},
  \bibinfo{person}{Dongyan Zhao}, {and} \bibinfo{person}{Rui Yan}.}
  \bibinfo{year}{2021}\natexlab{}.
\newblock \showarticletitle{Learning an effective context-response matching
  model with self-supervised tasks for retrieval-based dialogues}. In
  \bibinfo{booktitle}{\emph{Proceedings of the AAAI Conference on Artificial
  Intelligence}}, Vol.~\bibinfo{volume}{35}. \bibinfo{pages}{14158--14166}.
\newblock


\bibitem[Ye et~al\mbox{.}(2022)]%
        {GCN-Survey}
\bibfield{author}{\bibinfo{person}{Jiexia Ye}, \bibinfo{person}{Juanjuan Zhao},
  \bibinfo{person}{Kejiang Ye}, {and} \bibinfo{person}{Chengzhong Xu}.}
  \bibinfo{year}{2022}\natexlab{}.
\newblock \showarticletitle{How to Build a Graph-Based Deep Learning
  Architecture in Traffic Domain: A Survey}.
\newblock \bibinfo{journal}{\emph{IEEE Transactions on Intelligent
  Transportation Systems}} \bibinfo{volume}{23}, \bibinfo{number}{5}
  (\bibinfo{year}{2022}), \bibinfo{pages}{3904--3924}.
\newblock


\bibitem[You et~al\mbox{.}(2020)]%
        {18GraphCL}
\bibfield{author}{\bibinfo{person}{Yuning You}, \bibinfo{person}{Tianlong
  Chen}, \bibinfo{person}{Yongduo Sui}, \bibinfo{person}{Ting Chen},
  \bibinfo{person}{Zhangyang Wang}, {and} \bibinfo{person}{Yang Shen}.}
  \bibinfo{year}{2020}\natexlab{}.
\newblock \showarticletitle{Graph Contrastive Learning with Augmentations}. In
  \bibinfo{booktitle}{\emph{Proceedings of the 34th International Conference on
  Neural Information Processing Systems}} \emph{(\bibinfo{series}{NIPS'20})}.
\newblock


\bibitem[Yu et~al\mbox{.}(2022)]%
        {RGSL}
\bibfield{author}{\bibinfo{person}{Hongyuan Yu}, \bibinfo{person}{Ting Li},
  \bibinfo{person}{Weichen Yu}, \bibinfo{person}{Jianguo Li},
  \bibinfo{person}{Yan Huang}, \bibinfo{person}{Liang Wang}, {and}
  \bibinfo{person}{Alex Liu}.} \bibinfo{year}{2022}\natexlab{}.
\newblock \showarticletitle{Regularized Graph Structure Learning with Semantic
  Knowledge for Multi-variates Time-Series Forecasting}. In
  \bibinfo{booktitle}{\emph{Proceedings of the Thirty-First International Joint
  Conference on Artificial Intelligence, {IJCAI-22}}}.
  \bibinfo{pages}{2362--2368}.
\newblock


\bibitem[Yu et~al\mbox{.}(2021)]%
        {15recommend}
\bibfield{author}{\bibinfo{person}{Junliang Yu}, \bibinfo{person}{Hongzhi Yin},
  \bibinfo{person}{Min Gao}, \bibinfo{person}{Xin Xia},
  \bibinfo{person}{Xiangliang Zhang}, {and} \bibinfo{person}{Nguyen~Quoc
  Viet~Hung}.} \bibinfo{year}{2021}\natexlab{}.
\newblock \showarticletitle{Socially-aware self-supervised tri-training for
  recommendation}. In \bibinfo{booktitle}{\emph{Proceedings of the 27th ACM
  SIGKDD Conference on Knowledge Discovery \& Data Mining}}.
  \bibinfo{pages}{2084--2092}.
\newblock


\bibitem[Zhao et~al\mbox{.}(2022)]%
        {07ST-GSP}
\bibfield{author}{\bibinfo{person}{Liang Zhao}, \bibinfo{person}{Min Gao},
  {and} \bibinfo{person}{Zongwei Wang}.} \bibinfo{year}{2022}\natexlab{}.
\newblock \showarticletitle{ST-GSP: Spatial-Temporal Global Semantic
  Representation Learning for Urban Flow Prediction}. In
  \bibinfo{booktitle}{\emph{Proceedings of the Fifteenth ACM International
  Conference on Web Search and Data Mining}} \emph{(\bibinfo{series}{WSDM
  '22})}. \bibinfo{pages}{1443–1451}.
\newblock


\bibitem[Zhou et~al\mbox{.}(2021)]%
        {Informer}
\bibfield{author}{\bibinfo{person}{Haoyi Zhou}, \bibinfo{person}{Shanghang
  Zhang}, \bibinfo{person}{Jieqi Peng}, \bibinfo{person}{Shuai Zhang},
  \bibinfo{person}{Jianxin Li}, \bibinfo{person}{Hui Xiong}, {and}
  \bibinfo{person}{Wancai Zhang}.} \bibinfo{year}{2021}\natexlab{}.
\newblock \showarticletitle{Informer: Beyond efficient transformer for long
  sequence time-series forecasting}. In \bibinfo{booktitle}{\emph{Proceedings
  of the AAAI Conference on Artificial Intelligence}}.
  \bibinfo{pages}{11106--11115}.
\newblock


\bibitem[Zhou et~al\mbox{.}(2020)]%
        {Zhou2020}
\bibfield{author}{\bibinfo{person}{Zhengyang Zhou}, \bibinfo{person}{Yang
  Wang}, \bibinfo{person}{Xike Xie}, \bibinfo{person}{Lianliang Chen}, {and}
  \bibinfo{person}{Hengchang Liu}.} \bibinfo{year}{2020}\natexlab{}.
\newblock \showarticletitle{RiskOracle: A Minute-Level Citywide Traffic
  Accident Forecasting Framework}. In \bibinfo{booktitle}{\emph{Proceedings of
  the AAAI Conference on Artificial Intelligence}}.
  \bibinfo{pages}{1258--1265}.
\newblock


\end{thebibliography}


\clearpage
\appendix

\section{Spatial and Temporal Positional Embedding Method for Transformer Encoder}\label{appendA}
In our spatial-temporal synchronous contrastive learning stage, the module is built upon Transformer architecture, which employs self-attention mechanism as the key component. Previous works~\cite{Fourier-PE,Informer} have suggested that attention mechanism is agnostic to the sequential order, thus it's necessary to add proper positional embeddings as additional position information for Transformer. In our case, spatiotemporal traffic data has both temporal positional information (the traffic sequential data of each node individual comes from different historical time steps) and spatial positional information (at a given time step, different nodes in a traffic network has specific spatial distribution), the two kinds of positional information are both significant for accurate traffic forecasting. Thus, we explicitly introduce temporal positional embedding $PE_{T}$ and spatial positional embedding $PE_{S}$ to STS-CCL model, as described in the following.

\paragraph{Temporal Positional Embedding Method}
We adopt the commonly adopted fixed positional embedding method here. Suppose the input traffic data from time step $\tau$ and $j$-th feature dimension, its corresponding temporal positional embedding can be formulated as:
\begin{equation}
\begin{array}{lr}
PE_{T}(\tau,2j)=\sin[\tau/(2L_x)^{\frac{2j}{d_{model}}}], & \\
PE_{T}(\tau,2j+1)=\cos[\tau/(2L_x)^{\frac{2j}{d_{model}}}]
\end{array}
\label{eq:eqxx}
\end{equation}
\noindent where $1\leq j \leq d_{model}$, $d_{model}$ is the representation feature dimension, $L_x$ is the total temporal length of historical traffic data. 

\paragraph{Spatial Positional Embedding Method}
Spatial positional embedding requires us not only to represent each node's spatial position but also reflect the traffic graph structure. In considering of that, we first generate a geographic location embedding $Emb_{g}$ using sinusoid transformation and fully-connected network, then a GCN layer is used to obtain the final spatial positional embedding $PE_{S}$. Given a traffic network consisting of $N$ nodes, a geographic location matrix $G$ is created to represents the spatial coordinate information. Next, the sinusoidal transform $ST(G,\gamma_{\min},\gamma_{\max})$ is applied, where $\gamma_{\min},\gamma_{\max}$ denote the minimum and maximum grid scale, respectively. Following that, a fully-connected layer $FC_{\theta}(\cdot)$ is employed to reshape it into desired size $Emb_{g}$. The above procedure can be formulated as follows:
\begin{equation}
Emb_{g}=FC_{\theta}(ST(G,\gamma_{\min},\gamma_{\max}))
\label{eq:eqxxx}
\end{equation}

As suggested by~\cite{Li2018deeper}, GCN is a type of method to realize Laplacian smoothing, which could enhance the correlation representation between a node and its neighbors. Thus, we use GCN layer as the last process to generate the final spatial positional embedding $PE_{S}=GCN(Emb_{g})$.

\section{Implementation Details} \label{appendB} 
We split all the datasets into 60\%, 20\%, 20\% for training, validation and testing. All of our experiments are repeated for five times with different seeds, the mean and standard deviation values are also reported. For modeling training, the batch-size is set as 64, epoch is set as 100. Adam optimizer is adopted, with a learning rate of $1e^{-4}$ and weight decay of $1e^{-4}$. For data augmentation techniques, we evaluate the best ratio of Edge/Attribute masking in Fig.~\ref{fig4} and use the optimal settings for STS-CCL. In experiments, Attribute masking is directly implemented by masking the input traffic data, as suggested by~\cite{STGCL}. The proposed temporal scale fusion augmentation is input specific, where the hyper-params $\alpha,\beta$ are generated in every training epoch. In the overall objective function $L_{STS-CCL}$, the parameter $\epsilon$ is tuned within range $[0.1,1.0]$ with an increasing step of 0.1. We find the best $\epsilon$ is 0.70 for Hangzhou-Metro and 0.40 for Seattle-speed. In Transformer, the Encoder and Decoder layers are set as $N=4, N'=4$. For semantic contextual contrastive learning, the temperature parameter $\Delta$ is set as 0.1. The proposed STS-CCL and all other baseline methods are implemented with PyTorch 1.10 framework on two NVIDIA RTX-4090 GPU for neural computing acceleration.

\section{STS-CCL Model Training Algorithm}\label{appendC}
\begin{algorithm}[tbh]
\caption{The Model Training Algorithm of STS-CCL.}
\label{alg:alg1}
\begin{flushleft}
\textbf{Input}: Historical data $X$; ground-truth future traffic data $Y$; \\
Model parameter Top-$u$; $\Delta$; $\epsilon$; \\
Transformer encoder $E_{\theta}(\cdot)$; Transformer decoder $D_{\omega}(\cdot)$; \\
Non-linear projection head $g_{\phi}(\cdot)$ in semantic contextual contrasting; \\
Data augmentation method $\text{Aug}_s(),\text{Aug}_b()$; \\
The hard mutual-view prediction strategy $mvp()$; \\
The similar spatial heterogeneity filtering method $filter_{s}$; \\
The semantic context similarity filtering method $filter_{sem}$;\\
\textbf{Output}: $E_{\theta}(\cdot)$, $D_{\omega}(\cdot)$, $g_{\phi}(\cdot)$;\\
\end{flushleft}
\begin{algorithmic}[1]
\FOR {a batch of training sample $\{X_{i}\}_{i=1}^{N} \in X$}
\FOR {i=1 to N}
\STATE $z_i^{b}=\text{Aug}_b(X_i)$, $z_i^s=\text{Aug}_s(X_i)$.
\STATE $C_i^{b},C_i^{s}=E_{\theta}(z_i^{b}),E_{\theta}(z_i^{s})$.
\STATE ++++Stage one contrastive learning branch++++
\STATE $L_{sts}^b=mvp(C_i^{b}), L_{sts}^s=mvp(C_i^{s})$.
\STATE ++++Traffic prediction branch++++
\STATE $\hat{X}_i=D_{\omega}(C_i^{b})$.
\STATE ++++Stage two contrastive learning branch++++
\STATE $H_i^b=g_{\phi}(C_i^{b})$, $H_i^s=g_{\phi}(C_i^{s})$.
\STATE $\mathbb{N}_{i}\leftarrow filter_s(X_i), filter_{sem}(X_i,Top-u)$.
\ENDFOR
\STATE $L_{pred}=\frac{1}{N}\sum_{i=1}^N (Y_i-\hat{X}_i)^2$.
\STATE $L_{cl}=L_{sts}^{b}+L_{sts}^{s}+L_{sc}$.
\STATE $L_{sc}=L_{sc}=\frac{1}{N} =-\log \frac{\exp(sim(H_{i}^b,H_{i}^s)/\Delta)}{\sum_{j \in \mathbb{N}_{i}} \exp(sim(H_{i}^b,H_{j}^s)/\Delta)}$
\STATE $L_{sts}^s=-\frac{1}{N} \sum_{i=1}^{N} \log \frac{\exp ((W_i (C_{i}^s))^{\mathrm{T}}Z_{i}^b)}{\sum_{n \in \mathbb{N}_{i}} \exp (((W_i (C_{i}^s))^{\mathrm{T}})Z_n^b)}$;\\
\STATE $L_{sts}^b=-\frac{1}{N} \sum_{i=1}^{N} \log \frac{\exp ((W_i (C_{i}^b))^{\mathrm{T}}Z_{i}^s)}{\sum_{n \in \mathbb{N}_{i}} \exp((W_i(C_{i}^b))^{\mathrm{T}}Z_{n}^s)}$;\\
\STATE $L_{STS-CCL}=L_{pred}+\epsilon L_{cl}$.
\STATE Update $E_{\theta}(\cdot)$, $D_{\omega}(\cdot)$, $g_{\phi}(\cdot)$ to minimize $L_{STS-CCL}$.
\ENDFOR
\STATE \textbf{return}:$E_{\theta}(\cdot)$, $D_{\omega}(\cdot)$, $g_{\phi}(\cdot)$.
\end{algorithmic}
\end{algorithm}

\section{Contrastive Learning for Graph Structure Data} \label{appendD}
The underlying idea of contrastive learning is to achieve mutual recognition among different representations of the same data by using appropriate transformations. More recently, the huge research interest of contrastive learning has been extended from CV/NLP to spatial-temporal graph data. The purpose of spatial-temporal graph contrastive learning can be categorized into Node-level and Graph-level contrasting based on the specific downstream task. For instance, graph classification is a typical Node-level contrasting task, which is also the major research object of recently proposed contrastive models. Nevertheless, spatiotemporal contrastive learning for traffic forecasting task has not been deeply explored yet. 

The contrastive training procedure is briefly described as follows. In training batch $l$, a total number of $N$ node/graph individuals are processed by the contrastive neural network and resulting in $2N$
samples, where $N$ samples come from the Basic Augmentation View, another $N$ samples from the Strong Augmentation view. Let $s_i^{'}$ and $s_i^{''}$ represent the embedded features of node/graph $i$ from the basic and strong augmentation view, respectively. The basic graph contrastive loss equation is shown in Eq.~\ref{eq:eq2}:
\begin{equation}
\small
L_{gc}=\frac{1}{N} \sum_{i=1}^{N} -\log \frac{\exp(\text{sim}(s_i^{'},s_i^{''})/\sigma)}{\sum_{j=1,i \neq j}^{N} \exp(\text{sim}(s_i^{'},s_j^{''})/\sigma)}
\label{eq:eq2}
\end{equation}

\noindent where $\text{sim}(,)$ is the cosine similarity function, $\sigma$ is the temperature parameter. In graph contrastive loss $L_{gc}$, we can observe that there are $(N-1)$ number of negative pairs computed in the denominator. After pre-training, the learned spatiotemporal representations can be used to carry out specific tasks.

\end{document}